\RequirePackage{fix-cm}
\documentclass[smallcondensed]{svjour3}     %
\smartqed  %
\usepackage{graphicx}
\usepackage{amsmath} %
\usepackage{amssymb}  %

\usepackage{bm}  %

\usepackage[caption=false]{subfig}

\usepackage{siunitx}

\usepackage{algpseudocode} %
\usepackage{algorithm} %
\usepackage[T1]{fontenc} %
\usepackage{wrapfig} %

\usepackage[
        backend=biber,
        style=numeric,
        sorting=none,
        doi=false,
        isbn=false,
        url=false]{biblatex}

\newcommand{\dbm}[1]{\dot{\bm{#1}}}
\newcommand{\ddbm}[1]{\ddot{\bm{#1}}}

\newcommand{\eq}[1]{Eq.~(\ref{#1})}
\newcommand{\fig}[1]{Fig.~\ref{#1}}
\journalname{Journal for Intelligent \& Robotic Systems}
\begin{document}

\title{
        Virtual Forward Dynamics Models for Cartesian Robot Control
}

\author{Stefan Scherzinger         \and
        Arne Roennau \and R{\"u}diger Dillmann %
}

\institute{Stefan Scherzinger \\
        Arne Roennau \\
        R{\"u}diger Dillmann \at
              FZI Forschungszentrum Informatik \\
              Haid-und-Neu-Str 10-14 \\
              76131 Karlsruhe, Germany \\
              Tel.: +0049-721-9654-226\\
              \email{\{scherzinger, roennau, dillmann\}@fzi.de}           %
}

\date{Received: date / Accepted: date}

\maketitle

\begin{abstract}

In industrial context, admittance control represents an important scheme in
        programming robots for interaction tasks with their environments.
Those robots usually implement high-gain disturbance rejection on joint-level
        and hide direct access to the actuators behind velocity or position
        controlled interfaces.
Using wrist force-torque sensors to add compliance to these systems, force-resolved control laws
must map the control signals from Cartesian space
to joint motion.  Although forward dynamics algorithms would perfectly fit to
that task description, their application to Cartesian robot control is not well
researched.  This paper proposes a general concept of virtual forward dynamics models for Cartesian
robot control and investigates how the forward mapping behaves in comparison to
well-established alternatives.
Through decreasing the virtual system's link masses in comparison to the end effector, the
virtual system becomes linear in the operational space dynamics.
        Experiments focus on stability and manipulability,
particularly in singular configurations.
        Our results show that through this trick, forward
dynamics can combine both benefits of the Jacobian inverse and the Jacobian
transpose and, in this regard, outperforms the Damped Least Squares method.

\end{abstract}

\section{Introduction}
\label{introduction}
In robotics, task space control is important for many applications, since it
provides a natural way for programmers to specify goals and constraints.
The according control laws can be formulated in operational space of the end-effector.
Since the robots are articulated mechanisms and are powered in their joints, these controllers
need to map the Cartesian control signals to the robots'
configurations space, i.e. the motor actuators.
We will refer to matrices that accomplish this as \textit{mapping matrices}.
Two frequently-used variants of mapping matrices are the transpose of the
manipulator's Jacobian and its inverse.

The Jacobian transpose is an important part in many classes of control schemes
for torque-actuated robots, such as in
hybrid force/motion control~\cite{Whitney1977}, \cite{Raibert1981}, \cite{Khatib1987},
parallel force/motion control~\cite{Chiaverini1993},
and impedance control~\cite{Hogan1985}, \cite{Villani2008}.
In principal, the approaches use the Jacobian transpose as static relationship between
end-effector wrenches and joint torques for controlling robots in contact with their environments.
Although not strictly required, control performance is generally improved
through decoupling robot dynamics in operational space \cite{Khatib1987}, prior
to mapping the signals to joint space.
In addition, there are also algorithmic solutions using the principle of
\textit{inverse dynamics} to compute suitable joint torques from motion control signals, e.g.
with the Recursive Newton-Euler Algorithm (RNEA)~\cite{Featherstone2008}.

An important subset of robots, however, does not provide joint interfaces on torque level.
These systems are often found in industrial context, and are the primary focus
of this paper.
In \cite{Villani2008}, those systems are referred to as
\textit{simplified systems} because they hide internal dynamics decoupling behind a velocity interface.
In this paper, we will refer to them as \textit{velocity-actuated} systems to underline the type of interface exposed by the robot vendors.
On these systems, velocity-resolved control variants, such as admittance control~\cite{Villani2008},
usually leverage Jacobian inverse-related methods, such as the Damped Least Squares (DLS)~\cite{Wampler1986} as the mapping to joint space.

Unlike inverse dynamics for torque-actuated systems,
literature on velocity-actuated robots mostly neglects \textit{forward
dynamics} as an algorithmic option for control.
This is surprising, because it represents a straightforward mapping from
Cartesian wrench space to joint accelerations.
While we used this method to control robots in previous work
\cite{Scherzinger2017}, \cite{Scherzinger2019Inverse}, the new contribution of this paper is an in-depth analysis of
particular features of this mapping and an evaluation against other well-established methods.
The goal and novelty is a drop-in-replacement for the Jacobian inverse and DLS
in controllers for velocity-actuated robots.
Through using a dynamics-conditioned, virtual forward model, we match
the linear, decoupled behavior of the Jacobian inverse while simultaneously
keeping the inherent robustness in singularities of the Jacobian transpose
method.

The paper is structured as follows:
In \ref{sec:problem_statement} we briefly recapitulate the inverse kinematics mapping problem
along with established methods to make it easy for the reader to follow comparisons in the experiments.
Section \ref{sec:forward_dynamics} then presents forward dynamics-based mappings for Cartesian control.
In the experiments section \ref{sec:experiments}, we investigate ill conditioning, stability and manipulability in singular configurations and evaluate our approach against suitable subsets of the Jacobian Inverse, the Jacobian transpose and the DLS method.
We finally discuss remaining points and suggestions in
\ref{sec:discussion} and conclude with directions for further research in \ref{sec:conclusion}.

\section{Problem statement and related work}
\label{sec:problem_statement}

The goal of this paper is to evaluate forward dynamics-based mappings against 
well-established methods.
We use Singular Value Decomposition (SVD) as a tool to investigate
the characteristics of the mapping matrices.
SVD factorizes a matrix $\bm{M}$ according to
\begin{equation}
        \label{eq:singular_value_decomposition}
                \bm{M} = \bm{U} \bm{\Sigma} \bm{V}^T, \quad \text{with} ~\sigma_i = \Sigma_{ii}.
\end{equation}
$\bm{U}$ and $\bm{V}^T$ are orthogonal matrices. The entries $\sigma_i \geq 0$
of the diagonal matrix $\bm{\Sigma}$ are known as the singular values and
determine the scale of the mapping.
For our experiments, $\sigma_{\text{min}}$ and
$\sigma_{\text{max}}$ as the minimal and maximal singular values are
of particular interest in analyzing stability and manipulability.

To recapture some basic concepts, let the forward kinematics mapping be given with
\begin{equation}
        \label{eq:forward_kinematics}
                \bm{x} = g \left( \bm{q} \right),
\end{equation}
which computes an end-effector pose, denoted here with $\bm{x}$ from the joint state vector $\bm{q}$.
The velocity vector of generalized coordinates $\dot{\bm{q}}$ maps with
\begin{equation}
        \label{eq:manipulator_jacobian}
                \dot{\bm{x}} = \bm{J} \dot{\bm{q}}
\end{equation}
to end~effector velocity vector $\dot{\bm{x}}$, using the manipulator Jacobian
$\bm{J} = \bm{J} (\bm{q})$. We generally omit the joint vector dependency in
further notation for brevity reasons.
For non-redundant manipulators, the inverse mapping is given by
\begin{equation}
        \label{eq:manipulator_jacobian}
                \dot{\bm{q}} = \bm{J}^{-1} \dbm{x}~.
\end{equation}
Near singular configurations, $\bm{J}$ looses rank, such that its inverse becomes numerically unstable.
The respective mapping for end-effector forces and torques to joint space with
\begin{equation}
        \label{eq:force_joint_torque_relationship}
                \bm{\tau} = \bm{J}^T \bm{f}
\end{equation}
does not suffer from these instabilities.
However, $\bm{J}$ becoming rank-deficient means that some components of
$\bm{f}$ will lie in the nullspace of the Jacobian transpose, i.e. they will be
balanced by the mechanism's mechanics and will hence not be able to actuate the joints.
This effect is a severe limitation for controller implementations.

Applied to motion control, a formal investigation of the Jacobian transpose method and
a numerical solution to the Inverse Kinematics problem was presented
in~\cite{Wolovich1984}. The authors' solution derives from a simple, 2nd order dynamical system
\begin{equation}
        \label{eq:wolovich_dynamic_system}
                \ddot{\bm{q}} = \bm{K} \bm{J}^T  \left( \bm{x}^d - g(\bm{q}) \right) ~,
\end{equation}
computing joint accelerations from the difference of a desired pose $\bm{x}^d$ and the current pose as determined by the forward kinematics $g(\bm{q})$.
They show with a Lyapunov stability analysis that the system is asymptotically stable for
an arbitrary positive definite matrix $\bm{K}$.

Using $\bm{J}^T$ will serve as a lower bound and
baseline in stability considerations of our contribution.

The DLS method is an applications of the
Levenberg-Marquardt stabilization to manipulator control \cite{Nakamura1986},
\cite{Wampler1986} and tries to remove instabilities of $\bm{J}^{-1}$ near singular configurations.
Note that the original version as proposed in \cite{Wampler1986} uses partial velocity matrices, adding the benefit of allowing different reference frames for each element.
Since we don't make use of this feature, we replace it with the common
manipulator Jacobian $\bm{J}$ instead.
Similar to pseudo inverse methods for redundant systems, which minimize $\lVert \bm{J} \dot{\bm{q}} - \dot{\bm{x}} \rVert
^2$, the idea is to add a damping term $\alpha$ against excessive joint velocities
that will trade-off accuracy for stability near singular configurations with
\begin{equation}
        \label{eq:dls_approach}
        \lVert \bm{J} \dot{\bm{q}} - \dot{\bm{x}} \rVert ^2 + \alpha^2 \lVert \dot{\bm{q}} \rVert ^2.
\end{equation}
The solution that minimizes this quantity is given by
\begin{equation}
        \label{eq:dls_solution}
        \dot{\bm{q}} = ( \bm{J}^T \bm{J} + \alpha^2 \bm{I} )^{-1} \bm{J}^T \dot{\bm{x}},
\end{equation}
see e.g. \cite{Buss2004} for a derivation.
The matrix $(\bm{J}^T \bm{J} + \alpha^2 \bm{I})$ is non-singular, which can be
shown with singular value decomposition \cite{Buss2004} and hence is guaranteed to be invertible.
This method is well established for practical control implementations and can serve as a drop-in replacement for $\bm{J}^{-1}$ in control loops.
We use this method as a baseline to compare our new forward dynamics-based
method in terms of manipulability.

A popular enhancement to DLS, using this method, is Selectively Damped Least Squares (SDLS)
\cite{Buss2005}.
The method converges faster and circumvents to choose
a suitable $\alpha$ by introducing singular vector-specific damping terms of the singular
value decomposition of $\bm{J}$ at the expense of a higher runtime cost.

Other methods include the more recent Exponentially Damped Least Squares
(EDLS)~\cite{Carmichael2017}, which is a solution with the focus on physical
Human-Robot interaction (pHRI).  Although effectively avoiding elbow-lock and
wrist-lock among other common singular phenomena, it requires explicit, albeit
easy parameterization by the user.

Both the Jacobian inverse and the Jacobian transpose have strengths and shortcomings and
present mappings for physically different control spaces.
When used on velocity-actuated systems, the Jacobian inverse does not need
dynamic decoupling, but suffers from instability, which DLS effectively
mitigates at the expense of loosing accuracy with increased damping. The Jacobian transpose needs
dynamics decoupling in the controller, but offers inherent stability
near singular configurations.
A general incentive is to obtain the best of these corner cases.
This paper proposes and empirically evaluates forward dynamics as a suitable approach to
achieve this combination at the core of closed-loop control schemes.

\section{Cartesian control with forward dynamics}
\label{sec:forward_dynamics}

\subsection{Forward dynamics simulations}
\begin{figure}
        \centering
        \includegraphics[width=.40\textwidth]{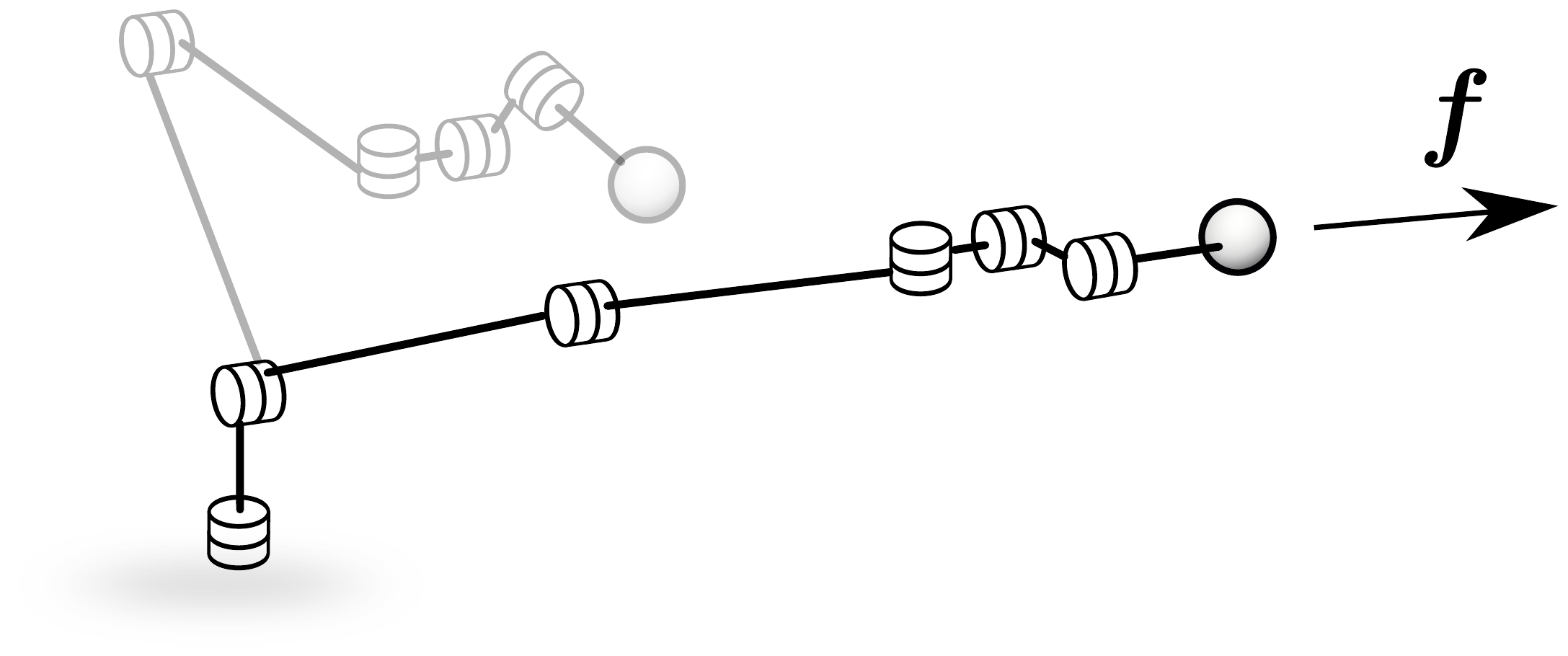}
        \caption{Illustration of pulling an exemplary robot manipulator into singularity.}
        \label{fig:forward_dynamics_simulation}
\end{figure}

To motivate the usage of forward dynamics in control applications let's
illustrate its behavior with a use case:
\fig{fig:forward_dynamics_simulation} depicts an arbitrary manipulator with joints and links.
Let's assume that the joints are pure articulations without motors and are
back-drivable, i.e. they can freely be moved.
An external force $\bm{f}$ is pulling the mechanism into singularity, in which the mechanism yields the external forces as good as it can, limited by kinematic constraints.
In the fully stretched case, increasing $\bm{f}$ will not create further motion.
The robot's mechanics compensate the external load until a possible breakage of the links.
This behavior is in fact inverse to how $\bm{J}^{-1}$ would compute joint
motion due to an external error vector, where theoretically infinite joint
velocity would occur.

To make use of forward dynamics simulations, robotic manipulators can be modeled as a system of articulated, rigid bodies.
The equations of motion describe the relationship
between generalized loads $\bm{\tau}$ in the joints, external loads $\bm{f}$, acting on the end-effector and motion in generalized coordinates $\bm{q}$ 
with the following ordinary differential equations in symbolic matrix notation
\begin{equation}
        \label{eq:rigid_body_system}
        \bm{\tau} + \bm{J}^T \bm{f} = \bm{H}({\bm{q}}) \ddot{\bm{q}} + \bm{C}( \bm{q},\dot{\bm{q}} ) + \bm{G}(\bm{q}) ~.
\end{equation}
$\bm{H}$ denotes the mechanism's positive definite inertia matrix,
$\bm{C}$ comprises the Coriolis and
centrifugal terms and $\bm{G}$ is the vector of gravitational components.
Forward dynamics computation has the goal of solving \eq{eq:rigid_body_system} for $\bm{q}(t)$, i.e.
simulating the mechanism's reaction motion through time, given external loads.

Literature has proposed several methods for forward dynamics computations, which can be categorized \cite{Featherstone2008} as mainly
belonging to inertia matrix methods with implementations of the
\textit{composite rigid body algorithm}, e.g. in \cite{McMillan1998},
\cite{Featherstone2005}, or the propagation methods with the
\textit{articulated body algorithm}~\cite{Featherstone1983} being an important
representative.
We refer the interested reader to \cite{Featherstone2008} for an broad coverage
of the field and recent implementation of various algorithms.

Forward dynamics is a substantial component in multi body simulations.
The fact that it is, however, mainly neglected for closed-loop control on
velocity-actuated systems may stem from the fact that computing good
approximations for $\bm{H}$ and $\bm{C}$ of the robots is
extremely difficult. The required crucial data, such as link masses and inertia
tensors is hardly available in data sheets.
On the other hand, a second reason for not being used might be that even if those data were
available, the benefit of forward simulating highly realistic motion would get
lost when executed on
velocity-actuated systems. Their internal joint servos with high-gain
disturbance rejection could not make use of the accuracy of dynamics that was
used to generate that motion.
The reference trajectory to follow would appear as any arbitrary trajectory.

This thought points to an interesting opportunity:
We could reduce \eq{eq:rigid_body_system} to a rough simplification and
investigate, if it's possible to condition $\bm{H}$ to beneficially \textit{tweak} the
behavior of this mapping when using it as a forward model in closed-loop
control. 
\subsection{A general closed-loop control}
\begin{figure}
        \centering
        \includegraphics[width=0.75\textwidth]{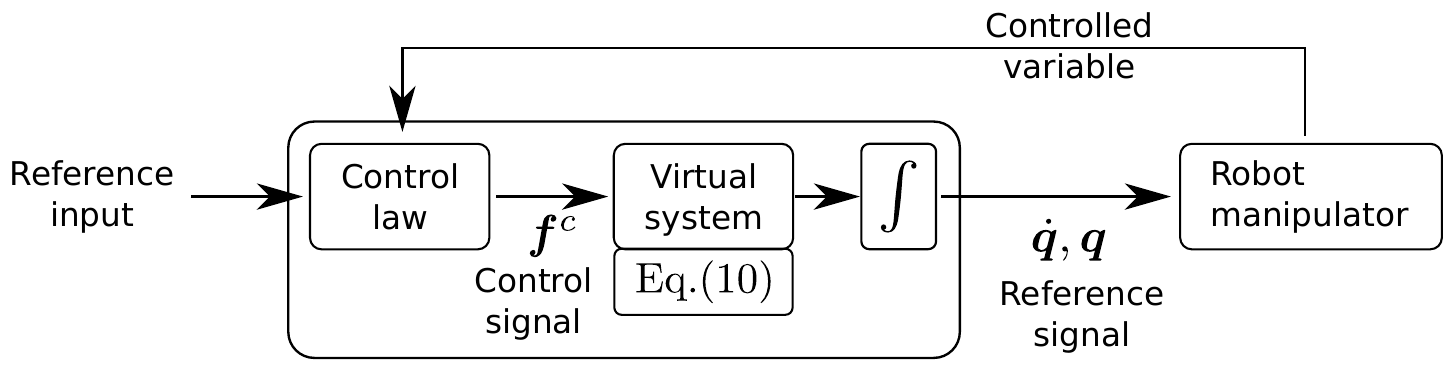}
        \caption{Closed-loop control with forward dynamics. We use a virtual system to forward simulate joint motion as a reference to the real robot.}
        \label{fig:closed_loop_control}
\end{figure}

To motivate simplifications to \eq{eq:rigid_body_system}, we investigate how a
controller would perceive the system in a possible closed loop control.
A general scheme is shown in \fig{fig:closed_loop_control}.
A suitable control law computes a Cartesian control signal $\bm{f}^c$, using a
user specified reference input and a controlled variable as feedback from the
robot.
Note the role of the \textit{virtual system} as a forward model in the controller:
We simulate how our proxy system behaves and send that as a reference to the real system.
Since we obtain joint accelerations as a response from our forward model, we
integrate those signals before sending them as reference to the real robot.
The advantage is, that this virtual model will react kinematically \textit{and}
mechanically plausible to external loads $\bm{f}^c$, as was illustrated in
\fig{fig:forward_dynamics_simulation}.

From the control law's point of view, a linear, virtual system would be beneficial
for using constant control gains for wide regions of the robot's joint configuration space.
By dropping the gravity term ($\bm{G}(\bm{q}) = \bm{0}$) from \eq{eq:rigid_body_system}, we assure that the control law does not need to constantly compensate this virtual load.
If we further consider instantaneous motion for each control cycle, i.e.
accelerate from rest with $\dot{\bm{q}} = \bm{0}$, we can drop the
non-linearities $\bm{C}( \bm{q},\dot{\bm{q}} )$ and obtain
\begin{equation}
        \label{eq:simplified_system}
        \ddot{\bm{q}} = \bm{H}^{-1}(\bm{q}) \bm{J}^{T} \bm{f}^c ~
\end{equation}
as an unbiased forward mapping.
We also set $\bm{\tau} \equiv \bm{0}$ to emphasize that $\bm{f}^c$ shall be
the only virtual load guiding the virtual system.

While dropping these terms reduces computational complexity in our controller,
including them can offer additional configuration.
This is briefly discussed in section \ref{sec:discussion}.

Note that $\bm{H}^{-1}(\bm{q})$ needs to be computed in each control cycle due
to its dependency of the current joint state. In the experiments section, we
evaluate computational cost in comparison to other mapping matrices.
Since \eq{eq:simplified_system} is effectively a Jacobian transpose-based
method, the next step is to decouple our virtual $\bm{H}^{-1}$.

\subsection{Decoupling virtual dynamics}
We start with the time derivative of \eq{eq:manipulator_jacobian}
\begin{equation}
        \label{eq:jacobian_diff}
                \ddot{\bm{x}} = \dot{\bm{J}} \dot{\bm{q}} + \bm{J} \ddot{\bm{q}}~~
\end{equation}
and consider instantaneous accelerations in each cycle while the virtual system is still at
rest, so that $\dot{\bm{J}} \dot{\bm{q}} = \bm{0}$.
With \eq{eq:simplified_system} we obtain
\begin{equation}
        \label{eq:cartesian_acceleration}
        \ddot{\bm{x}} = \bm{J} \bm{H}^{-1} \bm{J}^T \bm{f}^c = \bm{\Lambda}^{-1} \bm{f}^c~~,
\end{equation}
which describes the Cartesian instantaneous acceleration of the virtual system
due to the Cartesian control input $\bm{f}^c$.
The quantity $\bm{\Lambda}$ is known as the mass matrix in operational space,
see e.g. \cite{Khatib1987}, \cite{Villani2008}, with
$\bm{\Lambda} = \bm{J}^{-T} \bm{H} \bm{J}^{-1}$.

\begin{figure}
        \centering
        \includegraphics[width=0.30\textwidth]{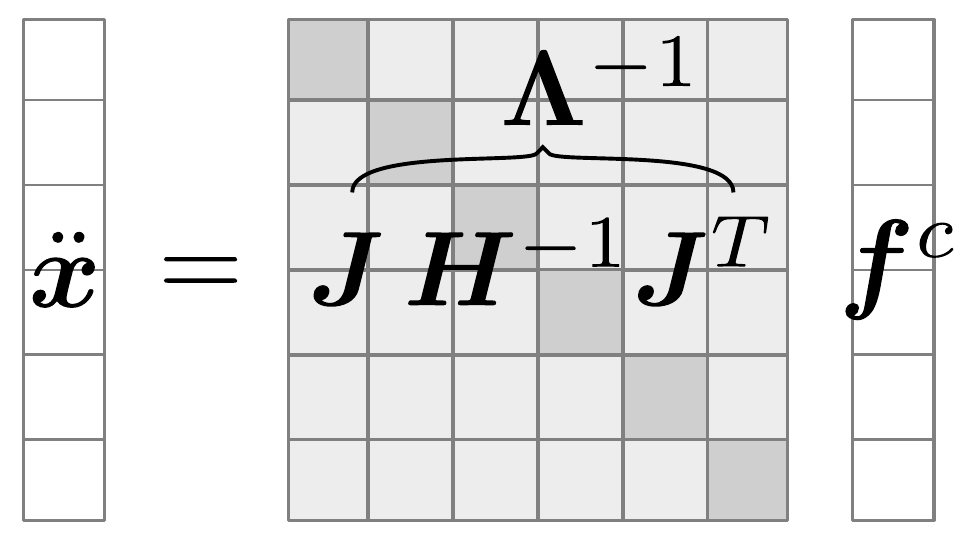}
        \caption{Graphical illustration of the mapping from wrench space to
        Cartesian acceleration. $\bm{\Lambda}$ is a $6 \times 6$ matrix for
        both redundant and non-redundant systems. Our goal is to obtain a
        decoupled, diagonal mapping for arbitrary joint configurations.}
        \label{fig:mapping_matrix_linearized}
\end{figure}

\begin{figure}
        \centering
        \includegraphics[width=.50\textwidth]{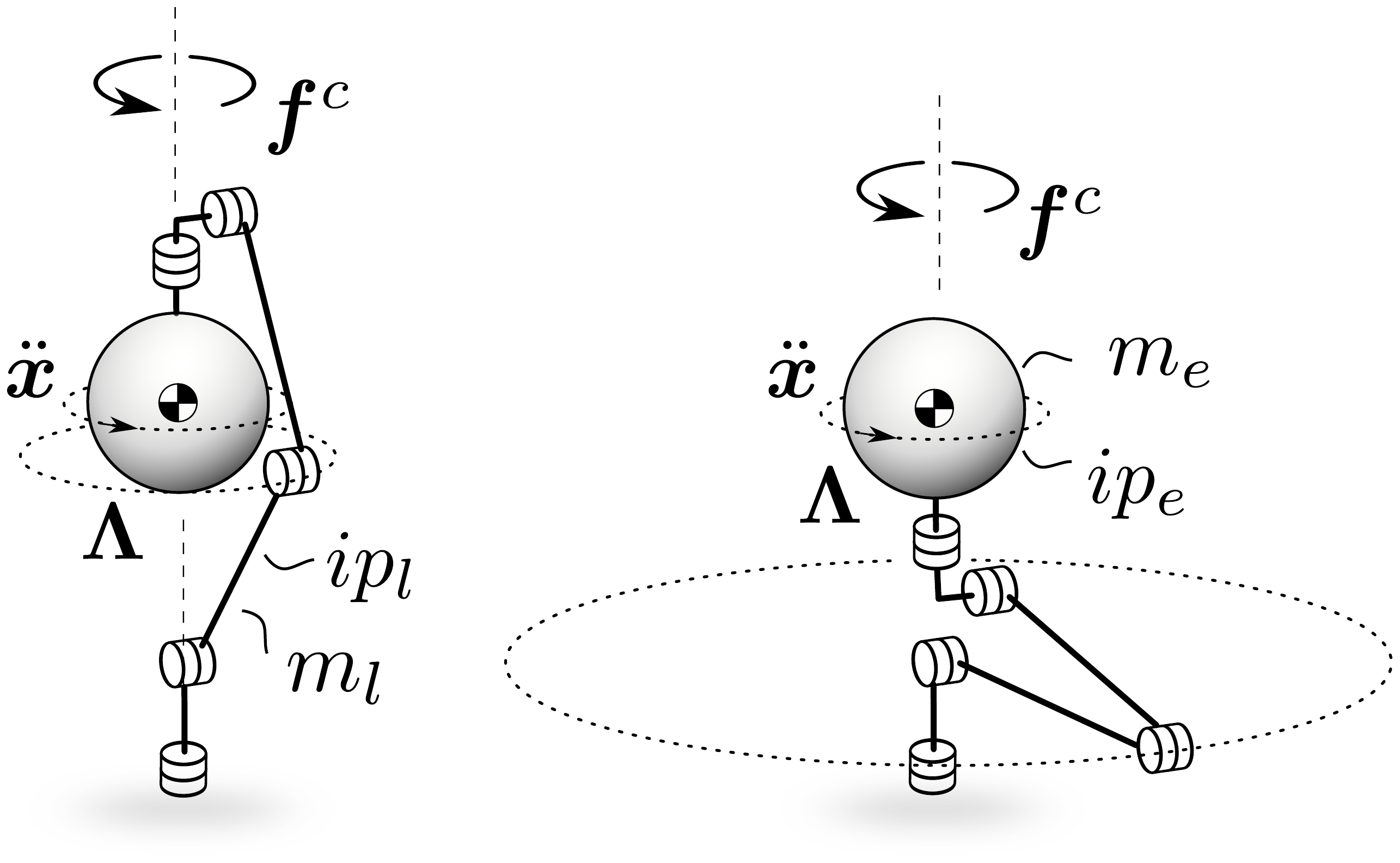}
        \caption{Dynamics-conditioned, virtual model for an exemplary robot.
        The goal is to make the mechanism behave as a unit mass, which is illustrated with the
        oversized sphere. }
        \label{fig:virtual_mechanism}
\end{figure}

The intention of our dynamic decoupling is to make $\bm{\Lambda}^{-1}$ a
time-invariant, diagonal matrix across joint configurations $\bm{q}$. This
ideal mapping is illustrated in \fig{fig:mapping_matrix_linearized}.
In order to preserve consistency of our virtual system and the real robot, we
use identical kinematics for both systems. This ensures that the reference
signals for the real robot to follow agree with possible limits.
We are, however, free in changing the dynamics of the virtual $\bm{H}$ to
obtain the desired effect, in particular its mass distribution.
The Cartesian control signal $\bm{f}^c$ acts directly on
the virtual mechanism's end~effector.
If that end-effector link is dominant with respect to the overall dynamics, determined by the other links,
we could obtain a behavior that comes close to an idealized unit mass.
\fig{fig:virtual_mechanism} illustrates this phenomenon.
As a consequence, the overall systems' center of mass roughly stays with the end-effector.
Likewise does the operational space inertia $\bm{\Lambda}$ depend less on joint configurations,
and $\bm{f}^c$ experiences the same rotational inertia for both configurations.
Having a realistic link mass distribution would instead mean higher inertia
with greater distance to the rotary axis.
To measure the effect of end-effector mass dominance, we define
\begin{equation}
        \label{eq:gamma}
        \gamma = \frac{m_e}{m_l} = \frac{ip_e}{ip_l}
\end{equation}
to be the ratio of end-effector mass $m_e$ and link mass $m_l$.
The quantities $ip_e$ and $ip_l$ denote the polar momentums of inertia of the
end-effector and the other links, respectively.
In the experiments section, we empirically show that increasing $\gamma$ in deed leads to
the desired behavior and provides a decoupled virtual system for Cartesian closed-loop control.

\subsection{Closed-loop stability}
Comparison of \eq{eq:wolovich_dynamic_system} and \eq{eq:simplified_system}
shows a strong resemblance of our forward dynamics-based approach with the
dynamical system from \cite{Wolovich1984}, if $\bm{f}^c = \bm{x}^d - g(\bm{q})$.
In \cite{Wolovich1984}, the authors prove
with a Lyapunov stability analysis
that a closed loop system, built from this mapping, is asymptotically stable for
an arbitrary positive definite matrix $\bm{K}$.
This formal proof also includes our proposed $\bm{H}^{-1}$, which is, due to
being grounded in the manipulator's kinetic energy $T = \dbm{q}^T \bm{H} \dbm{q}$, also positive definite.

\section{Experiments and results}
\label{sec:experiments}

We evaluated our forward dynamics-based approach against the DLS and against the two corner cases
$\bm{J}^{-1}$ and $\bm{J}^T$ in various experiments.
We chose the Universal Robot UR10's kinematics for our experiments.
Our perception is that this robot is well-known and used both in industry and academia and therefore presents a suitable platform.

\begin{figure}
        \centering
        \includegraphics[width=.60\textwidth]{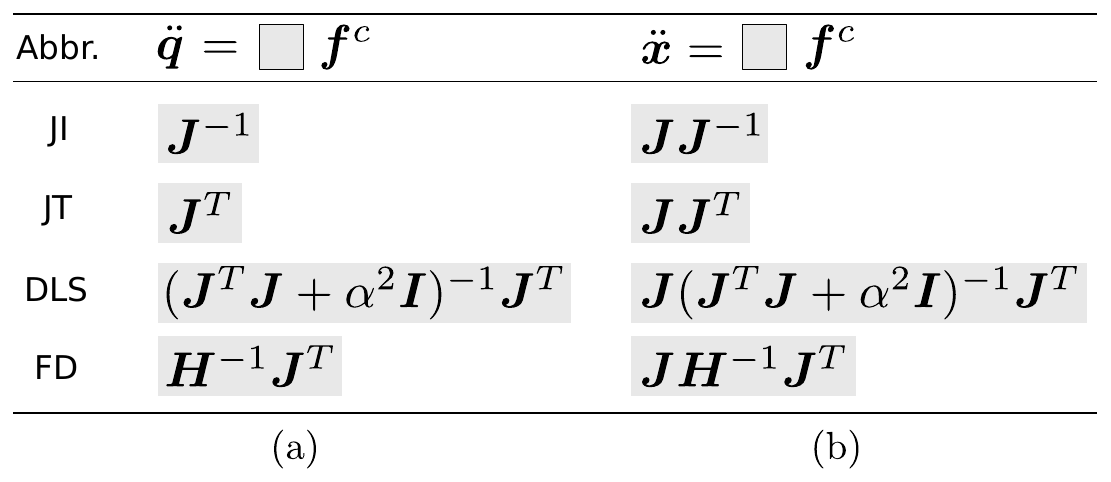}
        \caption{Mapping matrices for the experiments. The abbreviations stand
        for Jacobian inverse (JI), Jacobian transpose (JT), Damped Least
        Squares (DLS) and Forward Dynamics (FD). Two types are investigated:
        (a) Mappings from Cartesian space to joint space and (b) mappings from
        Cartesian space to Cartesian space.}
        \label{fig:mapping_matrices_experiments}
\end{figure}

Depending on the phenomena investigated, a subset of different mapping matrices was used.
An overview of these matrices and their composition is given in
\fig{fig:mapping_matrices_experiments} along with the abbreviations used in the
plots.
We implemented each mapping matrix literally, i.e. as a multiplication of the
respective symbols in C++, using the robot's kinematics from a popular
ROS~\cite{Quigley2009}
package\footnote{https://github.com/ros-industrial/universal\_robot} and the
algorithms for computing $\bm{J}$ and $\bm{H}$ from a well-established robotics
library\footnote{https://github.com/orocos/orocos\_kinematics\_dynamics}.

For all experiments, the following values were chosen for the forward dynamics mappings:
\begin{equation}
        \label{eq:parameters_for_experiments}
        m_e = \SI{1}{kg} ,\quad m_l = \frac{m_e}{\gamma}, \quad ip_e = \SI{1}{kg/m^2}, \quad ip_l = \frac{ip_e}{\gamma}
\end{equation}
The ratio $\gamma$ was then varied according to the investigation of each experiment.

\subsection{Decoupling}
\label{sec:decoupling}

In this experiment, we evaluated the effectiveness of our virtual model dynamics decoupling and compared the mapping to both the Jacobian inverse and the Jacobian transpose as reference.
The mapping matrices were of type (b) from \fig{fig:mapping_matrices_experiments}.
\begin{figure}
        \centering
                \includegraphics[width=0.8\textwidth]{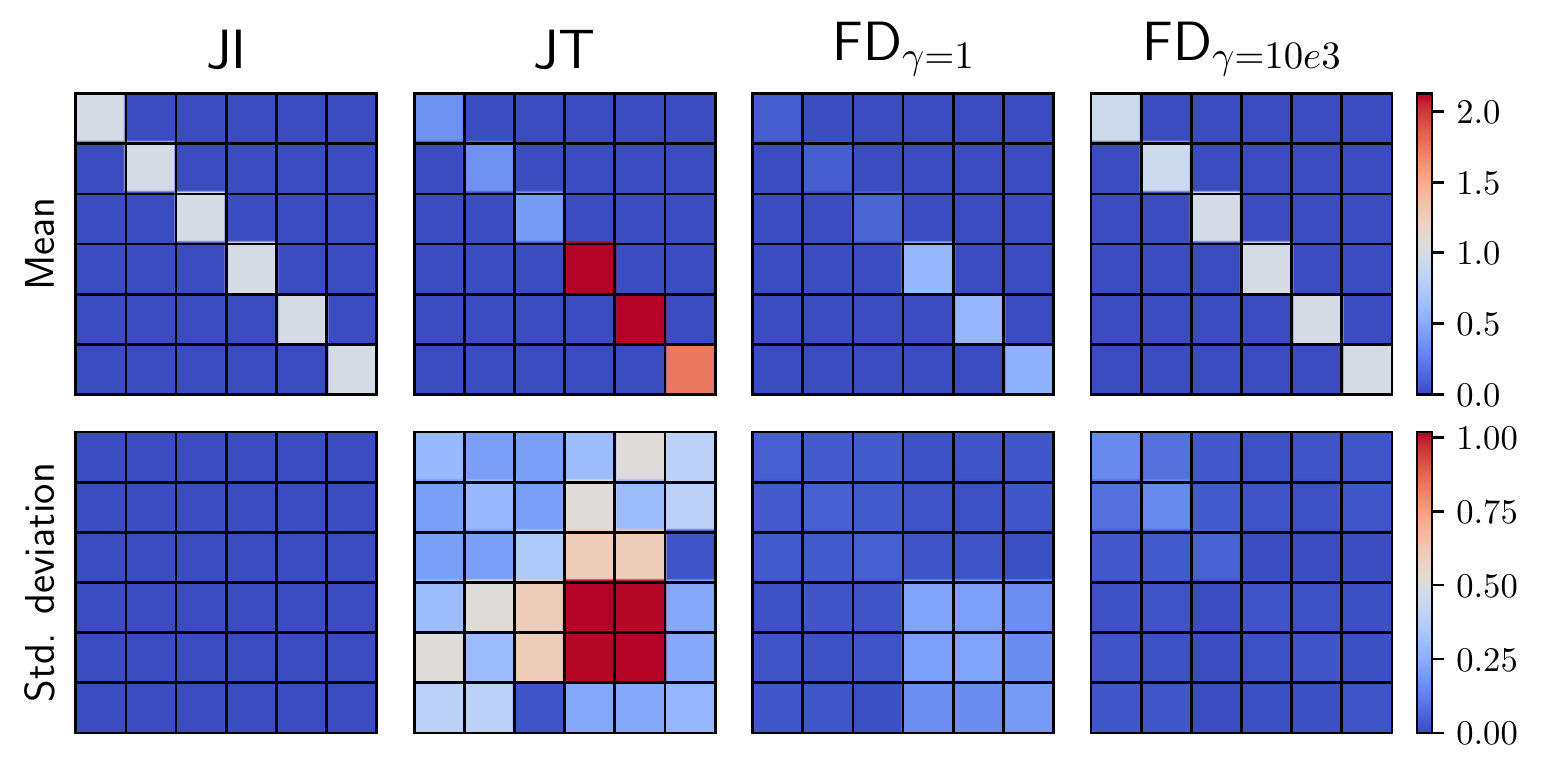}
        \caption{%
                Analysis of the $6\times6$ mapping matrices of type (b) from
                \fig{fig:mapping_matrices_experiments}.
                The plot shows the individual matrix entries in form of 2d heat plots.
                To obtain the plots, we sampled $100.000$ random configurations of $\bm{q}$
                uniformly with $q_i \in \left[-\pi, \pi \right]$ and computed
                the respective mapping matrix for each type.
                The figures show mean and standard deviation for the entirety of samples.
        }
        \label{fig:homogenization}
\end{figure}
\fig{fig:homogenization} shows the results of the analysis.
All mean matrices are diagonal, which is to expect for sampling a vast amount of arbitrary joint configurations.
The standard deviations, however, show a strong configuration dependency for the
Jacobian transpose.
This mapping would be suboptimal if used in a closed-loop control scheme.
Instead, the Jacobian inverse behaves ideal and converges to the identity matrix.
Note that using forward dynamics with an even mass distribution ($\gamma = 1$)
already improves upon the Jacobian transpose.
The experiment further shows, that with a significant end-effector mass
dominance ($\gamma = 10e^3$) the forward dynamics mapping converges to the
Jacobian inverse and makes this mapping particularly suitable for closed-loop
control in terms of linearity.

\subsection{Ill-conditioned configurations}
\begin{figure}
        \centering
        \includegraphics[width=.90\textwidth]{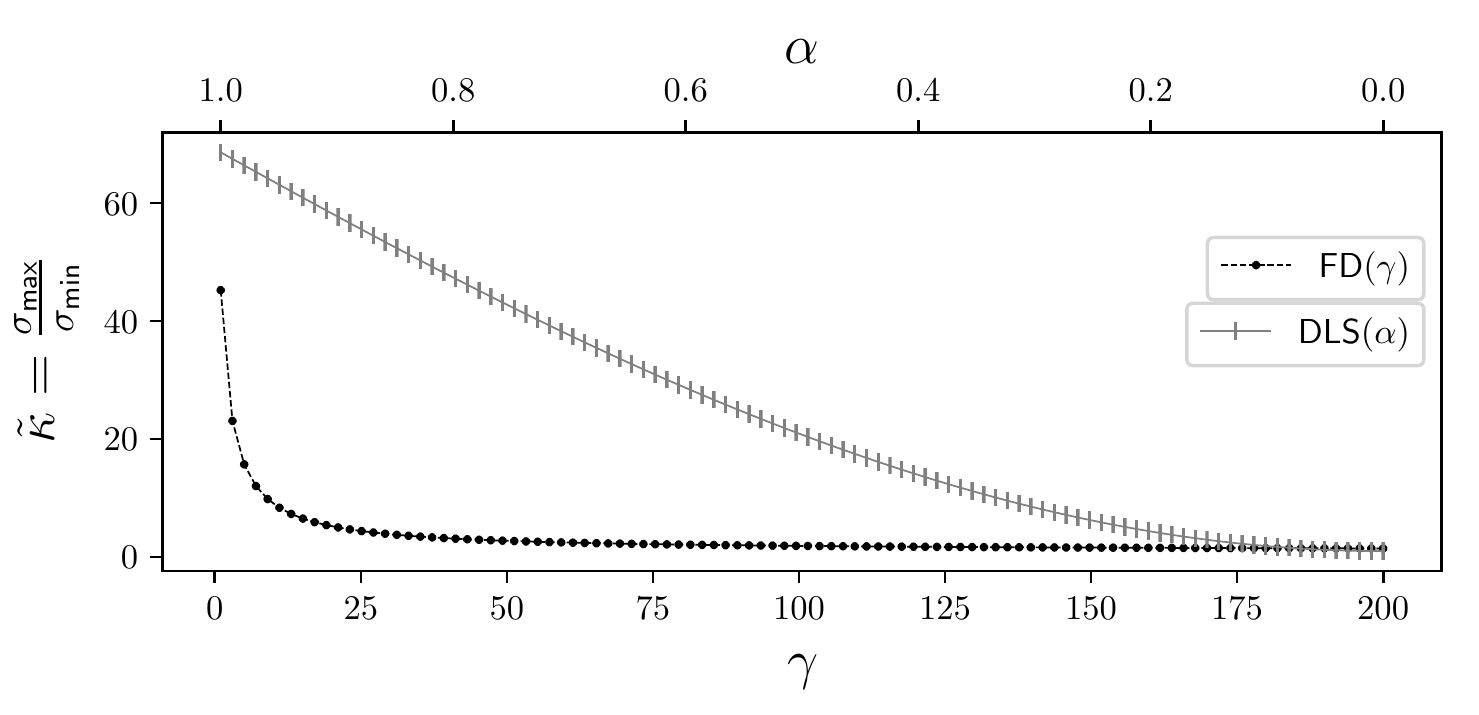}
        \caption{Ill conditioning $\kappa$ for the DLS and FD method. Each data
        point is the median of $1000$ randomly evaluated joint configurations}
        \label{fig:ill_conditioning}
\end{figure}

In this experiment, we continued the evaluation of decoupling and compared FD and DLS with regard to ill conditioning
of the mapping matrices from \fig{fig:mapping_matrices_experiments}(b).
Higher numbers of ill conditioning degrade control performance
\cite{Featherstone2004}, but heavily depend on the manipulators configuration.
This experiment investigates how FD and DLS influence ill conditioning
by varying $\gamma$ and $\alpha$, respectively.
Based on \cite{Featherstone2004}, we used $\kappa = \sigma_{\text{max}} /
\sigma_{\text{min}}$ as the measure for ill conditioning.
\fig{fig:ill_conditioning} shows the results.
For each discrete point in the plots, we evaluated $1000$ random joint states
with the limits from \fig{fig:homogenization}.
We used quartiles on our data to effectively exclude outliers ($ \sigma_{\text{min}} \rightarrow 0$), such that
the plots show the median of the ill conditioning.
It can be seen that FD converges much faster to beneficial condition numbers over its own parameter space than DLS.
In fact, most of the decoupling effect from experiment \fig{fig:homogenization}
is already available for low values of $\gamma$.
\subsection{Behavior in singularities}
\begin{figure}
        \centering
        \includegraphics[width=1.0\textwidth]{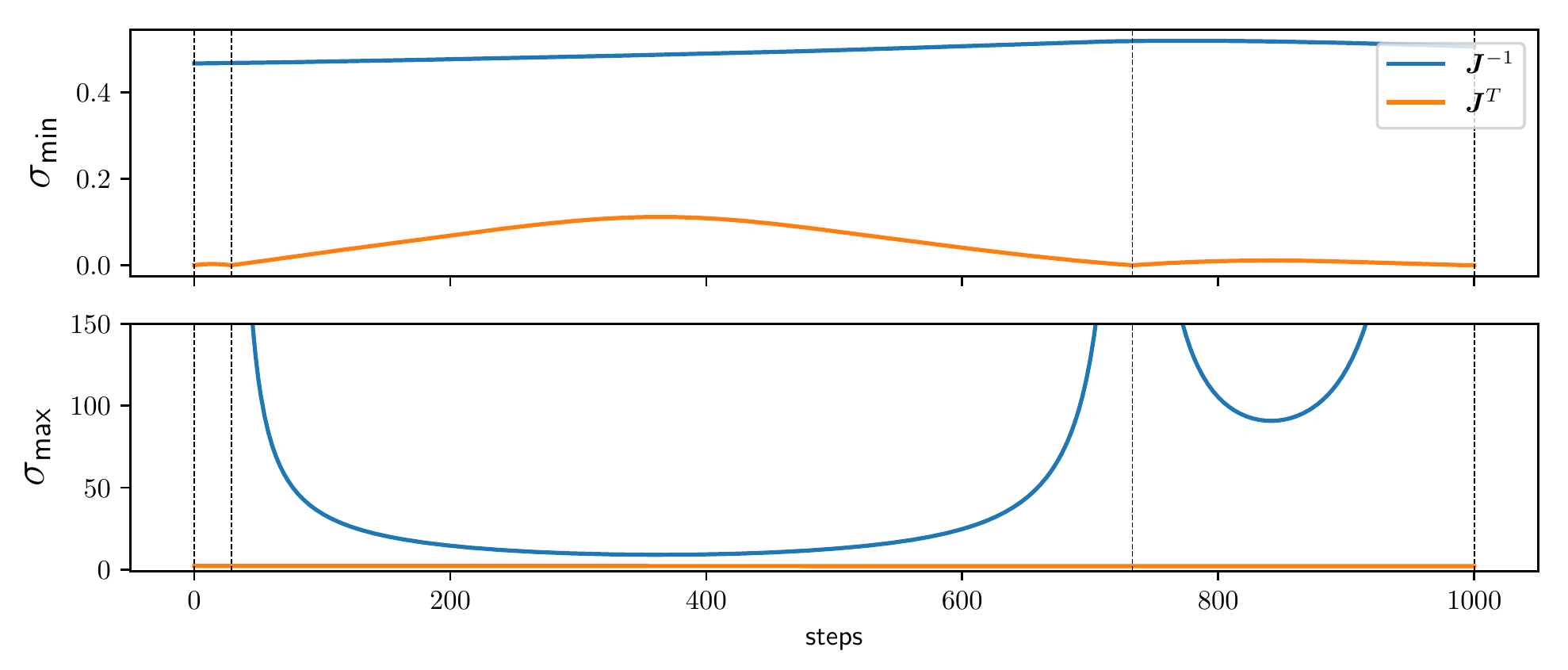}
        \caption{The two baselines $\bm{J}^T$ and $\bm{J}^{-1}$ while passing
        through four singular configurations, illustrated with dashed, vertical
        lines.}
        \label{fig:singular_configurations_baselines}
\end{figure}

Before reporting on this experiment, we briefly recapitulate desired and expected behavior in singularities.
In singular configurations, the manipulator Jacobian $\bm{J}$ becomes rank-deficient.
This is an unfortunate joint configurations for all considered approaches.
As a consequence, the manipulator is not able to achieve instantaneous motion in all directions~\cite{Murray1994}.
Two issues arise from this constellation:
First, Jacobian transpose-based methods tend to loose manipulability.
We measure this effect with $\sigma_{\text{min}}$ of the mapping matrix, which
is one of various established measures~\cite{Murray1994}.
Second, for Jacobian inverse-based methods, infinite joint velocity occurs.
We measure this effect with $\sigma_{\text{max}}$ as an indicator of how much
the mapping matrix scales $\bm{f}^c$ in sensitive dimensions to joint space.
The goal of the experiment is to investigate how well
each approach behaves in singular configurations concerning both measures.
As a reference, \fig{fig:singular_configurations_baselines} shows both the Jacobian inverse and
the Jacobian transpose for a pass through four singular configurations, the first two being close together.

The curves show the expected and well-known effect: The Jacobian inverse maintains high
manipulability at the cost of an exploding $\sigma_{\text{max}}$, while the
Jacobian transpose stays stable throughout the pass but cannot avoid
$\sigma_{\text{min}}$ dropping to zero in singularities.

\begin{figure}
        \centering
        \includegraphics[width=1.0\textwidth]{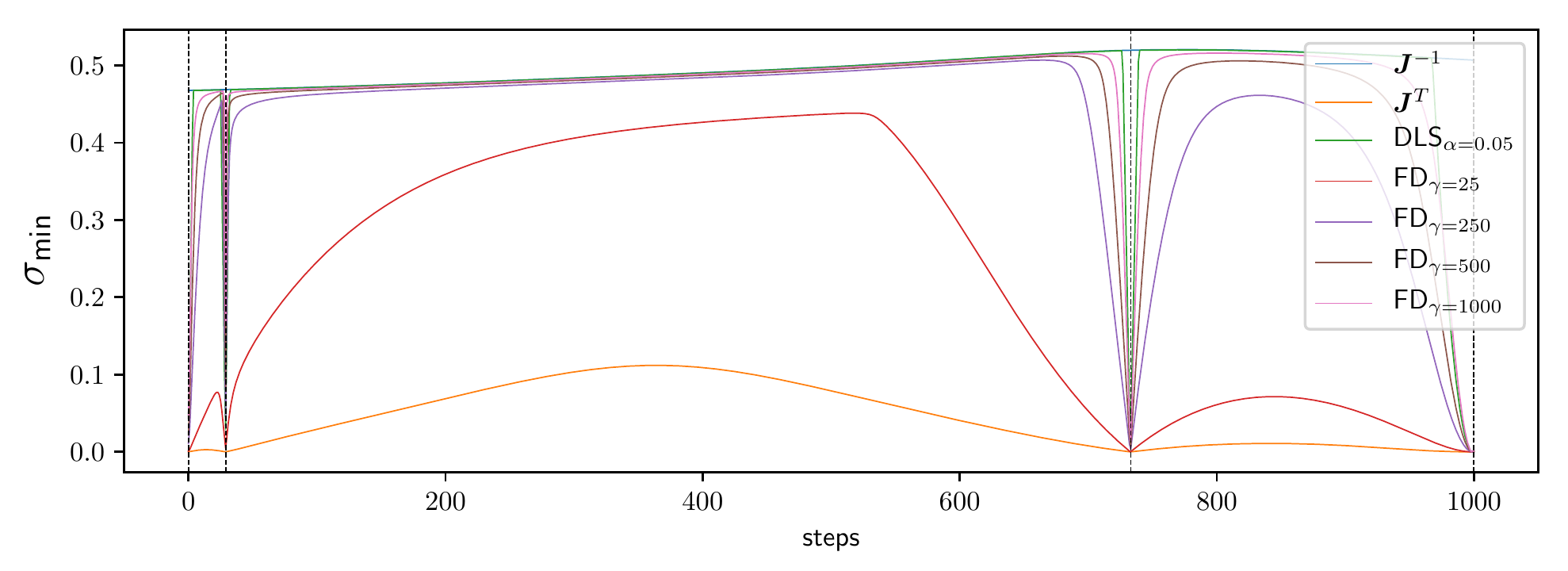}
        \caption{Investigation of $\sigma_{\text{min}}$ of various mapping matrices through four singular configurations.}
        \label{fig:singular_configurations_sigma_min}
\end{figure}

\begin{figure}
        \centering
        \includegraphics[width=1.0\textwidth]{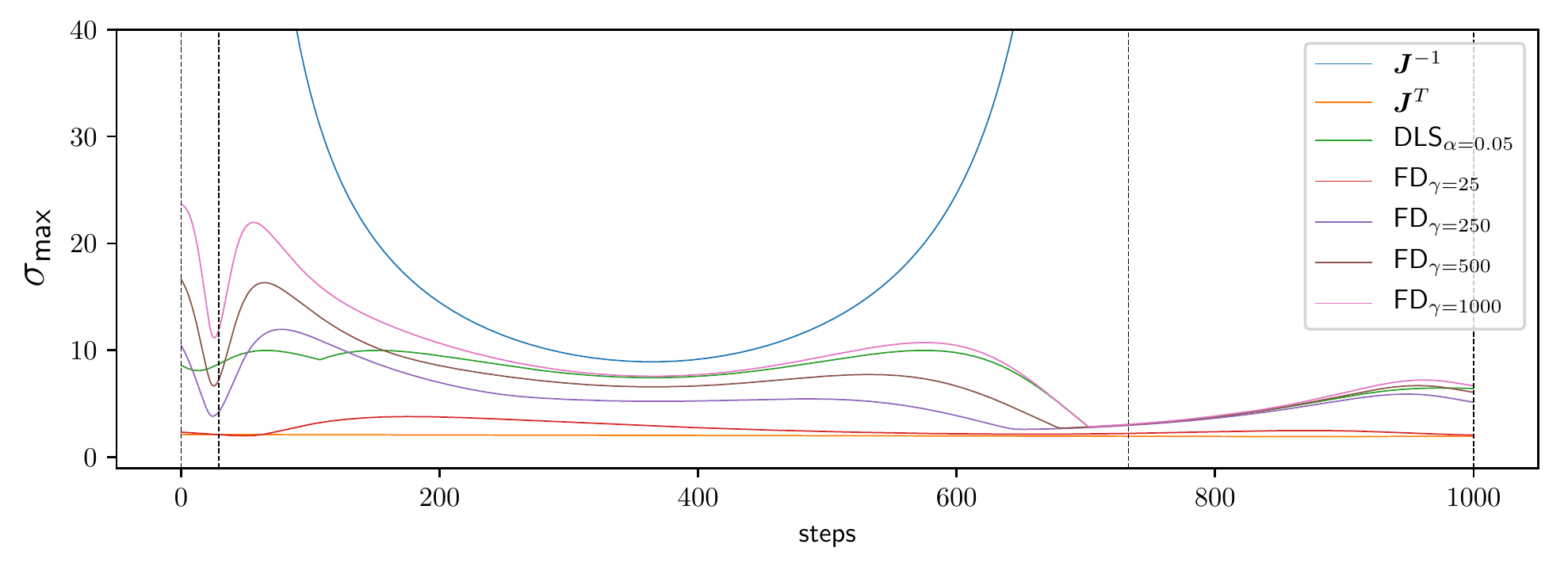}
        \caption{Investigation of $\sigma_{\text{max}}$ of various mapping matrices through four singular configurations.}
        \label{fig:singular_configurations_sigma_max}
\end{figure}

\fig{fig:singular_configurations_sigma_min} and
\fig{fig:singular_configurations_sigma_max} finally show the behavior of
forward dynamics with a set of different $\gamma$.
The curves show how FD approaches the Jacobian inverse while maintaining
$\sigma_{\text{max}}$ in stable ranges.
We added the DLS method, albeit with only one $\alpha$, for comparison.
Both FD and DLS have very similar characteristics and show a good trade-off
between both corner cases (JI, and JT).
Note how the curves for FD become more pronounced towards the Jacobian inverse for increasing $\gamma$.

\subsection{Empirical analysis of stability and manipulability}

In this experiment, we wanted to analyze FD in comparison to DLS on a broader scale.
The goal is an empirical analysis of varying $\alpha$ (DLS) and $\gamma$ (FD) over bigger ranges,
and evaluate how they perform in the corner cases in comparison to the Jacobian inverse and transpose.
Instead of focusing a few trajectories, we sampled a massive amount of singular constellations.
Note that in contrast to the workspace sampling for experiment
\ref{sec:decoupling}, which contained singular
configurations by chance, in this investigation we exclusively used singular
configurations.
Through exclusively focusing regions of low performance (singularities) throughout the
whole workspace, the results become a feasible measure of global performance for each method.

To find a big amount of singular configurations,
we first sampled equally distributed, random joint states as start configurations.
We then used Particle Swarm Optimization (PSO) \cite{Miranda2018PySwarms},
which implements an adapted algorithm from the original work of
\cite{Kennedy1995} as a black-box optimizing strategy to converge to singular
configurations from these start states.
We used Yoshikawa's manipulability measure $\sqrt{ \text{det} (\bm{J} \bm{J}^T)
}$, which simplifies for non-redundant mechanisms to $\lvert \text{det}
(\bm{J}) \rvert$ \cite{Yoshikawa1985Manipulability} as function to minimize,
which was faster than using SVD with $\sigma_{\text{min}}$ directly.
Alternatively, a more type-based approach of finding singularities is discussed
in \cite{Zlatanov1998}, \cite{Bohigas2012Numerical}.

\begin{figure}
        \centering
        \includegraphics[width=.90\textwidth]{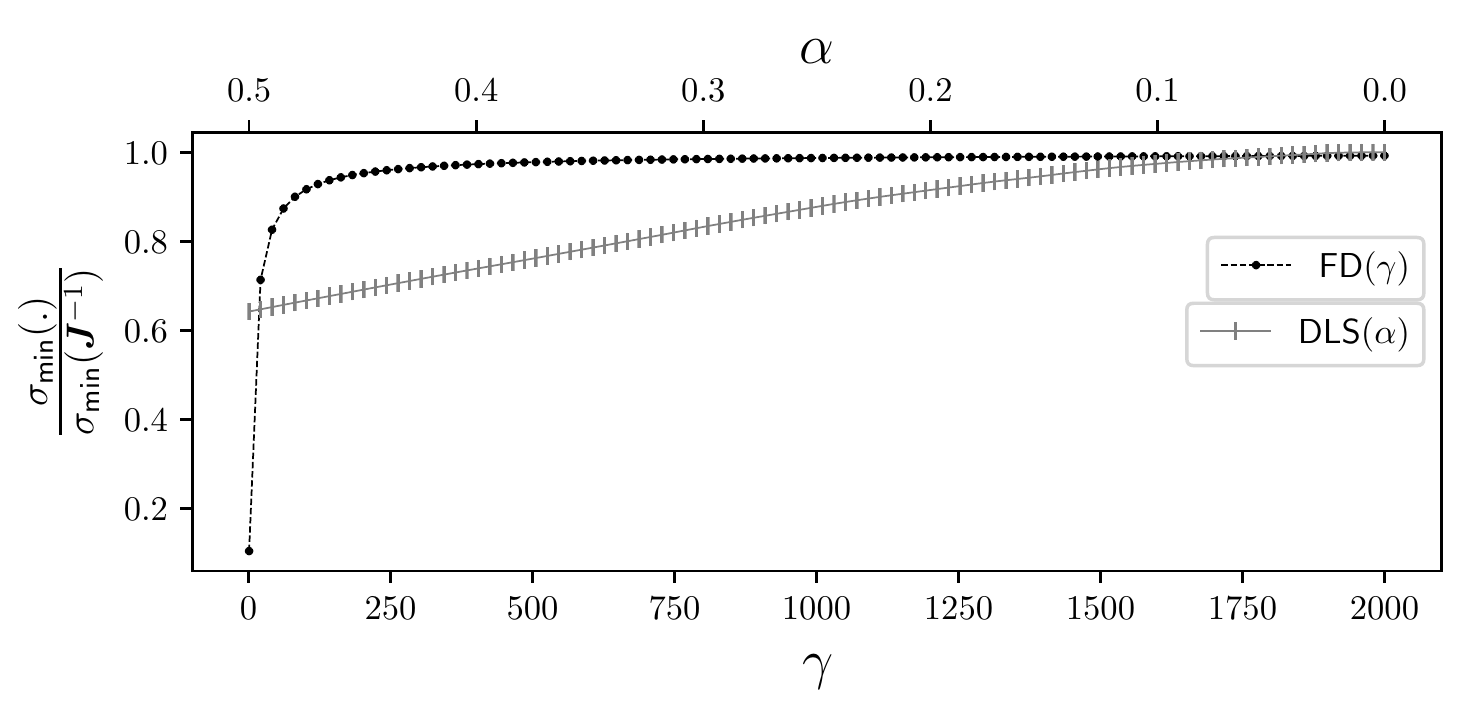}
        \caption{Relative manipulability for the DLS and FD method in comparison to $\bm{J}^{-1}$. Note, that practical
        applications of the DLS method may require higher damping values of up
        to $\alpha = 1.1$ as reported in \cite{Buss2005}}
        \label{fig:manipulability}
\end{figure}

Having a set of $1000$ singular configurations, we then computed average values
for $\sigma_{\text{min}}$ and $\sigma_{\text{max}}$ from the mapping matrices
of FD and DLS according to \fig{fig:mapping_matrices_experiments}(a) for
discrete values of $\alpha$ and $\gamma$ for each of the $1000$ singular
configurations.
\fig{fig:manipulability} shows the results for manipulability.
Both DLS and FD approach the Jacobian inverses behavior for decreasing $\alpha$ and increasing $\gamma$, respectively.
Note how FD approaches qualitatively faster in its own parameter space.

\begin{figure}
        \centering
        \includegraphics[width=.90\textwidth]{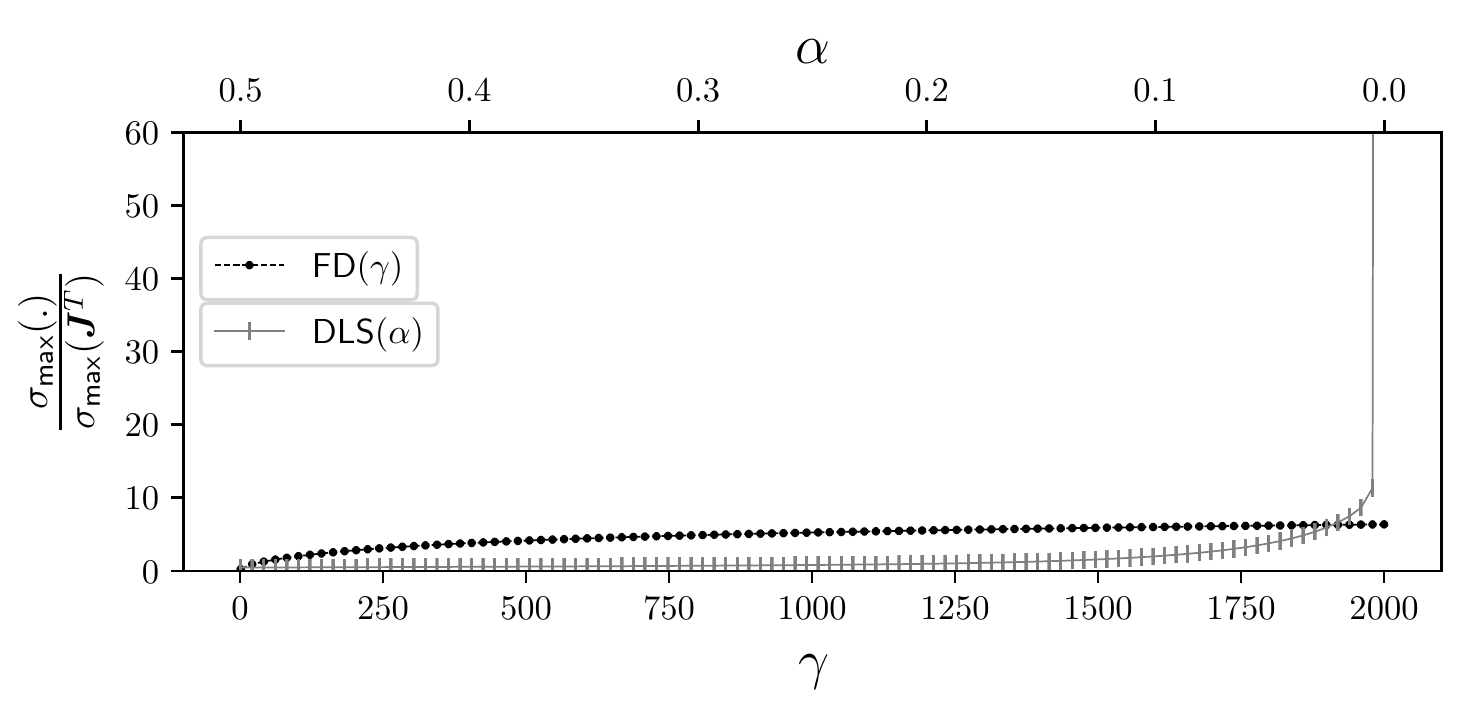}
        \caption{Relative instability for the DLS and FD method in comparison to $\bm{J}^T$.}
        \label{fig:stability}
\end{figure}

\fig{fig:stability} shows the results for stability.
DLS comes closer to the Jacobian transposes stability than FD throughout most
of the observed parameter space.
However, towards reaching the Jacobian inverses high manipulability, DLS
looses stability and asymptotically approaches infinity, while FD in contrast
stays bounded.
For applications in which DLS would require very low values of $\alpha$ for
control performance, FD can be used as a safe alternative, combining
\textit{and keeping} the benefits of both $\bm{J}^{-1}$ and $\bm{J}^T$.

\subsection{Computational efficiency}

Finally, we measured average execution times of the mapping approaches.
We implemented the SDLS method according to \cite{Buss2005} and included the
measurements as additional reference.

\begin{figure}
        \centering
        \includegraphics[width=0.9\textwidth]{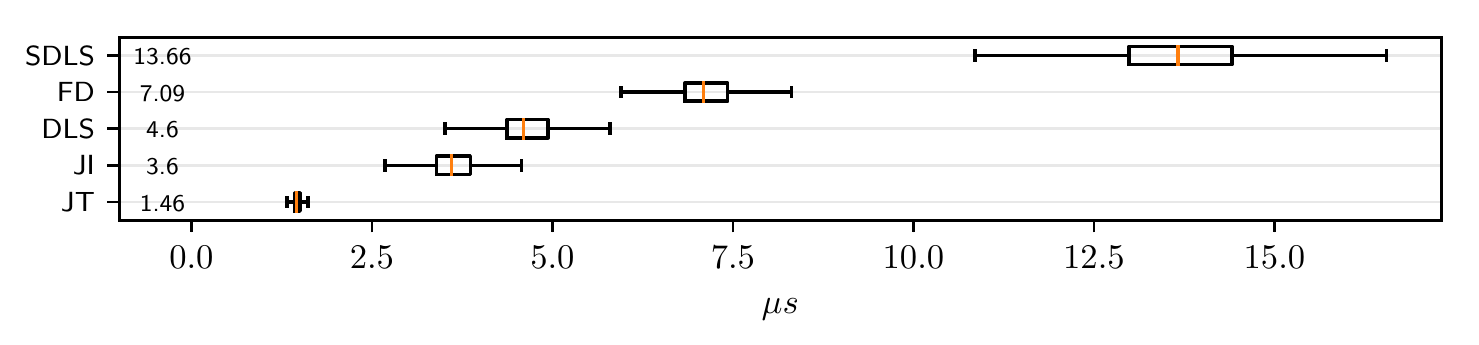}
        \caption{Execution times of computing different mappings of type
        (a) from \fig{fig:mapping_matrices_experiments} on an
        Intel\textsuperscript{\textregistered} Core\texttrademark~i7-4900MQ.
        }
        \label{fig:evaluate_performance}
\end{figure}

To obtain the comparison, we computed $\ddbm{q}$ as given in
\fig{fig:mapping_matrices_experiments}(a) with a fictitious, constant
$\bm{f}^c = \bm{1}$ for $10e5$ times.
The joint state $\bm{q}$ was randomly sampled, while being identical across one single
evaluation of each method.
\fig{fig:evaluate_performance} shows the boxplots of each method's execution
time with their quartiles. The median is plotted as vertical, orange line.
The whiskers for minimal an maximal execution times indicate a high degree of irregularity.
We expect narrower ranges for experiments on a hard real-time operating system.

The results show that the forward dynamics method is a little more
computationally intense than the DLS method, but approximately half the
execution time of the SDLS.
Being still in the low $\mu$s range makes forward dynamics in the version from this
paper suitable for real-time closed-loop control.

\section{Discussion}
\label{sec:discussion}
\subsection{Virtual forward models}
When deriving our principal mapping for forward dynamics in \eq{eq:simplified_system},
we dropped gravity and non-linear terms to support dynamics decoupling of our virtual system.
For redundant manipulators, including those terms offers additional interfaces
to adjust behavior in the nullspace of the Jacobian transpose.
For those cases, switching from the composite rigid body algorithm to
propagation methods for solving the forward dynamics might be beneficial.
The articulated body algorithm \cite{Featherstone1983}, e.g. allows an
intuitive integration of external loads to each link of the robot separately,
which might be used to implement collision avoidance or other optimizations
concerning the robots' posture.

\subsection{Control applications}
The natural mapping of forward dynamics from Cartesian wrench space to joint
accelerations makes it particularly suitable for the implementation of
admittance-related controllers on velocity-actuated systems.
For those use cases, force-resolved control laws for disturbance rejection can
replace the velocity-resolved control laws using DLS.
The benefit of using the FD method is its ability to operate extremely
close to the ideal $\bm{J}^{-1}$ behavior without significantly sacrificing
stability.
Successful implementations of forward dynamics-based control on industrial
robots can be found e.g. in \cite{Scherzinger2019Contact} for pure force control and in
\cite{Scherzinger2017}, \cite{Heppner2020} for compliance control.
An application to motion control with a particular focus on sparsely sampled
targets is presented in \cite{Scherzinger2019Inverse}.

\section{Conclusion}
\label{sec:conclusion}

This paper proposed virtual forward dynamics models for Cartesian robot control.
The core of the control loop is a simplified, virtual
model that maps Cartesian control signals to joint accelerations.
Through increasing the end effector's mass in comparison to the other links, the
virtual system becomes linear in the operational space dynamics and matches
the exactness of the Jacobian inverse.
Further experiments showed, that this forward model's decoupling
leads to less ill conditioning compared to the DLS method for an empirical
investigation of the joint space.
When passing through singularities, forward dynamics behaves in general similar
to DLS in terms of manipulability and stability.
Yet, when operating in singular configurations
forward dynamics models substantially differ from DLS in that they produce
bounded control signals, even when forced to approach the Jacobian inverse in
terms of manipulability.
These virtual forward models are particularly suitable for implementing
admittance controllers in industrial settings on velocity-actuated robots.
Their computation time in the low $\mu$s range makes them suitable for real-time control.

\printbibliography

\end{document}